# DETECTION OF SUBCLINICAL ATHEROSCLEROSIS BY IMAGE-BASED DEEP LEARNING ON CHEST X-RAY

Running title: The AI-CAC model for subclinical atherosclerosis detection

Guglielmo Gallone MD[1], Francesco Iodice[2], Alberto Presta PhD[2], Davide Tore MD[3], Ovidio de Filippo MD[1], Michele Visciano MD[2], Carlo Alberto Barbano PhD[2], Alessandro Serafini MD[3], Paola Gorrini[1], Alessandro Bruno[1], Walter Grosso Marra MD[4], James Hughes PhD[5], Mario Iannaccone MD[6], Paolo Fonio Prof[3], Attilio Fiandrotti Prof[2], Alessandro Depaoli MD[7], Marco Grangetto Prof[2], Gaetano Maria de Ferrari Prof[1], Fabrizio D'Ascenzo MD, PhD[1]

1. Division of Cardiology, Cardiovascular and Thoracic Department, Città della Salute e della Scienza Hospital, Turin, Italy; Department of Medical Sciences, University of Turin, Turin, Italy.
2. Computer Science Department, University of Turin, Turin, Italy
3. Radiology Unit, Department of Surgical Sciences, University of Turin, Città della Salute e della Scienza, Turin, Italy
4. Cardiology Division, Civil Hospital, Ivrea, Italy
5. Candiolo Cancer Institute, FPO-IRCCS, Candiolo, TO, Italy
6. Division of Cardiology, Ospedale San Giovanni Bosco, Turin, Italy
7. Radiology Division, Civil Hospital, Ivrea, Italy

**Corresponding author**

Guglielmo Gallone, MD

Division of Cardiology, Cardiovascular and Thoracic Department, Città della Salute e della Scienza Hospital, Turin, Italy; Department of Medical Sciences, University of Turin, Turin, Italy

Corso Bramante 88/90, 10126, Turin, Italy

Email: guglielmo.gallone@gmail.com Phone: +390116335443

**Word Count** (including references): 4501

# ABSTRACT

**Aims.** To develop a deep-learning based system for recognition of subclinical atherosclerosis on a plain frontal chest x-ray.

**Methods and Results.** A deep-learning algorithm to predict coronary artery calcium (CAC) score (the AI-CAC model) was developed on 460 chest x-ray (80% training cohort, 20% internal validation cohort) of primary prevention patients (58.4% male, median age 63 [51-74] years) with available paired chest x-ray and chest computed tomography (CT) indicated for any clinical reason and performed within 3 months. The CAC score calculated on chest CT was used as ground truth. The model was validated on an temporally-independent cohort of 90 patients from the same institution (external validation). The diagnostic accuracy of the AI-CAC model assessed by the area under the curve (AUC) was the primary outcome. Overall, median AI-CAC score was 35 (0-388) and 28.9% patients had no AI-CAC. AUC of the AI-CAC model to identify a CAC >0 was 0.90 (95%CI 0.84–0.97) in the internal validation cohort and 0.77 (95%CI 0.67–0.86) in the external validation cohort. Sensitivity was consistently above 92% in both cohorts. In the overall cohort (n=540), among patients with AI-CAC=0, a single ASCVD event occurred, after 4.3 years. Patients with AI-CAC>0 had significantly higher Kaplan Meier estimates for ASCVD events (13.5% vs. 3.4%, log-rank=0.013).

**Conclusion.** The AI-CAC model seems to accurately detect subclinical atherosclerosis on chest x-ray with elevated sensitivity, and to predict ASCVD events with elevated negative predictive value. Adoption of the AI-CAC model to refine CV risk stratification or as an opportunistic screening tool requires prospective evaluation.

## ABBREVIATIONS

ASCVD = atherosclerotic cardiovascular disease

AUC = area under the curve

CAC = coronary artery calcium

CAD = coronary artery disease

CT = computed tomography

ESC = European Society of Cardiology

HR = hazard ratio

IQR = interquartile range

# INTRODUCTION

Coronary artery disease (CAD) affects more than 34 million people in the European Union, representing the leading cause of death, with an estimated total healthcare cost of 18.8 billion/year(1).

Primary prevention measures and treatments are effective in reducing disease burden and limiting the adverse consequences of a cardiovascular event(2). Implementation on a societal scale of these costly strategies requires precise identification of patients at risk for a cardiovascular event to tailor clinical management. However, the ability to stratify the individual atherosclerotic cardiovascular disease (ASCVD) risk is still limited, hampering implementation of actions directed at high-risk individuals, both in terms of efficacy and cost-effectiveness. ASCVD risk stratification is traditionally based on the individual risk profile, including dyslipidemia, smoking, hypertension, diabetes, life-style and family history. However, this strategy lacks both sensitivity and specificity for ASCVD, leading to both under- and over-treatment(3–6).

In recent years, the progress in cardiovascular imaging modalities has provided the tools for accurate detection of subclinical coronary atherosclerosis(4,7), facilitating new risk stratification strategies focused on the presence of subclinical disease rather than on disease probability (i.e. the clinical risk profile). Specifically, subclinical coronary atherosclerosis can be precisely assessed by the presence of coronary artery calcium (CAC)(8). CAC testing has convincingly demonstrated to strongly predict ASCVD in asymptomatic individuals independently and more accurately as compared to a clinical profile-guided ASCVD risk stratification(9,10), resulting in a more effective guide for initiation or deferral of preventive therapies (3,6,11), translating into valuable clinical efficacy and cost-effectiveness(6,8,12,13).

A multidetector non-contrast enhanced computed tomography (CT) of the chest is currently needed to measure CAC. While recommended by most international clinical guidelines, CAC score remains underused in clinical practice as an effect of physician's uneasiness to prescribe a chest CT scan in asymptomatic individuals, the radiobiological risk, the costs associated with the exam, and the limited availability of this exam in some settings. With this background, the detection of subclinical atherosclerosis on a readily available, rapid, low-radiation, inexpensive exam such as chest radiography might overcome the limits of CT scan, prompting wide-spread utilization of a disease-based strategy to prevent ASCVD among asymptomatic individuals at risk.

Deep-learning and neural network, representing the highest degree of complexity in artificial intelligence technologies, may learn complex hierarchical representations buried into many hidden layers of information, to produce refined clusters of data from sources (images) whose information does not seem to be accessible to the human eye(14).

Thus, we sought to develop a deep-learning based system for recognition of subclinical atherosclerosis on a plain chest radiography, with the aim to facilitate wide-spread implementation of subclinical atherosclerosis-based risk stratification strategies.

## METHODS

### Study design and imaging data curation

The electronic archives of *Città della Salute e della Scienza Hospital* were searched for all chest x-rays and plain chest CT. When a chest x-ray and a chest CT of an individual patient indicated for any clinical reason were performed within 3 months, image data were extracted. All non-contrast chest CT with slice thickness below 2.5 mm were considered for inclusion. Clinical data regarding demographics, cardiovascular risk factors, cardiovascular pathology and cardiovascular events were extracted from the electronic health records up to last available follow-up. Clinical and imaging data were anonymized. Patients were excluded if having ascertained coronary events prior to the date of the chest x-ray, if not discharged alive (if the radiology exam was taken during a hospitalization), if deemed to have a non-cardiovascular condition with a prognosis <1 year, if chest x-ray was performed bed-side or in any other position different from the standard standing one, or if the chest CT image quality was deemed inadequate to assess CAC score from the radiologist. CAC score was calculated by two radiologists with more than 5-year experience in CT scan interpretation.

Cardiovascular (CV) risk was calculated according to the European Society of Cardiology (ESC) classification into low, intermediate, high and very high CV risk classes(2,15).

### AI-CAC model development

The aim was to develop a deep-learning based system (the AI-CAC model) for recognition of subclinical atherosclerosis on a plain frontal chest x-ray. To develop the AI-CAC model, the CAC score assessed on chest CT was used as ground truth label (i.e. gold standard against whom the model was trained). The full methodology to develop the AI-CAC model is described in the Supplementary methods (Supplementary Appendix). Briefly, following pre-processing of the chest x-ray images (Figure 1), a neural network consisting of a deep convolutional feature extractor followed by one prediction head for CAC score prediction was developed. Concerning the features extractor, the approach(16) relying on a DenseNet121(17) backbone was followed. It is composed of 120 convolutional layers attached to a prediction head to obtain the predicted value of CAC. The output of the DenseNet feature extractor is a vector of 1024 features maps that is the representation of the input image. Concerning the prediction head, it consists of a fully connected hidden layer with 64 neurons with ReLu activation function(18) and one output layer. The network yields for each input a real-valued number representing the CAC score, so the output layer includes only one neuron with linear activation.

**AI-CAC model training and validation.**

As a first step, the feature extractor was trained on the Chexpert dataset(19), representing one of the largest chest X-rays datasets with a total of 224,316 images from 65,240 patients, following the same supervised procedure previously described(16). Next, the derivation cohort was randomly split into two datasets: a training (80%) cohort, which was used to train the AI-CAC model, and an internal validation (20%) cohort, which was used to test the developed model on unseen data. A retrospective independent population was also collected and analyzed to further test the AI-CAC model in an external validation cohort, pertaining to a subsequent historical period at the same institution, during which partial renovation of the chest x-ray machines occurred. To explore model generalizability, a sensitivity analysis was carried out to explore the model performance in the external validation subset of chest x-rays acquired with renovated machine models, under-represented in the derivation cohort.

**Outcomes.**

The diagnostic accuracy of the AI-CAC model as compared to CAC score was the primary study outcome. The prognostic performance of the AI-CAC model to predict the composite of ASCVD (comprising acute coronary syndrome, cerebrovascular event, coronary revascularization, coronary death and sudden cardiac death) up to 5 years was the secondary outcome. The clinical outcome was assessed in the pooled study cohorts.

Sensitivity analyses were performed to assess the diagnostic and prognostic performance of the AI-CAC model according to demographics, CV risk factors and ESC CV risk classes.

**AI-CAC model explainability.**

In order to provide explainability for the AI-CAC model, a method called GradCam, also defined as Gradient-weighted class activation mapping, was used. This is a technique widely exploited to interpret, or at least justify, the decisions made by a convolutional neural network, and has the main goal to find what parts of the image are more relevant to make a decision. In particular, it leverages the gradient of the loss function with respect to the features maps of the last convolutional layer to extract an activation map which highlights the parts of the image that contributed the most to the model's decision.

**Statistical analysis**

Categorical variables are expressed as number and percentages, continuous variables are expressed as mean ± standard deviation or median and interquartile range (IQR) as appropriate. Unpaired t test or nonparametric Mann-Whitney U test were used for comparisons of continuous variables and chi-square test was used for categorical variables.

The diagnostic accuracy of the AI-CAC model considered as a continuous variable to predict a CAC >0 was assessed in terms of area under the receiver operating characteristic curve (AUC). The sensitivity, specificity and predictive values of an AI-CAC value >0 to predict a CAC value >0 were also assessed.

Kaplan-Meier survival curves and log-rank p values were performed along with univariate and multivariate Cox regression analyses to assess the association of AI-CAC with ASCVD events in a time-dependent fashion, and its independence from the ESC CV risk grading. Results are presented as hazard ratio (HR) and 95% confidence intervals (CIs).

A $p <0.05$ was considered statistically significant. Statistical analyses were conducted using SPSS (version 24.0, SPSS Inc., Chicago, Illinois, US) and STATA (version 17, StataCorp, College Station, Texas).

## RESULTS

### Study population

Of 700 patients with paired CT scan and chest X-ray, 92 patients were excluded for coronary events history, 80 patients for being affected by a non-cardiovascular condition with a prognosis <1 year or because non discharged alive, and 68 for chest CT image quality that was deemed inadequate to assess CAC score from the radiologist, leading to a derivation cohort of 460 patients. The external validation cohort comprised 90 patients.

Demographics, CAC score and CV risk profile data overall and stratified by no AI-CAC vs. AI-CAC >0 are reported in Table 1 for the pooled cohorts and in Supplementary Table 3 (Supplementary Appendix) for the derivation and external validation cohorts.

514 (93.5%) patients had complete information on CV risk profile and 535 (97.3%) patients had complete ASCVD events assessment qualifying for the respective analyses.

Overall, 58.4% of patients were male and median age was 63(51-74), 27.2% of patients had 0 CV risk factors, 57.5% 1-2 CV risk factors and 15.5% >2 CV risk factors. 4.1% were at low, 45.3% at intermediate, 26.3% at high and 24.3% at very high CV risk according to the ESC CV risk stratification classification. Overall, median AI-CAC score was 35(IQR 0-388) and 28.9% patient had no AI-CAC. Patients with no AI-CAC score were younger, more frequently female and had a lower CV risk profile and lower CAC levels. Overall, patients in the external validation cohort were younger, with a lower burden of CV risk profile, and with lower CAC (Supplementary Table 3).

### AI-CAC diagnostic accuracy

The discriminative performance of the AI-CAC model to identify a CAC >0 as expressed by the AUCs are shown in **Figure 1**. AUCs were 0.90 (95%CI 0.84–0.97) in the internal validation cohort and 0.77 (95%CI 0.67–0.86) in the external validation cohort, respectively. Precision-recall curves are provided in Supplementary Figure 3.

The accuracy metrics for an AI-CAC value >0 to detect a CAC >0 are reported in **Table 2**. Sensitivity was consistently above 92% across both cohorts while specificity was modest to moderate (between 69.4% at internal validation and 45.5% at external validation).

The distribution of X-ray machine models adopted for image acquisition were similar between the internal validation and the training cohorts (p-trend=0.566), while significantly different between the external validation and the training cohorts (p<0.001) (Supplementary Figure 4). Specifically, machine models 3 and 5 (external validation cohort vs. training cohort: 48.9% vs. 12.5%, p<0.001) were under-represented in the derivation cohort and increasingly used in the external validation cohort. In this subset, the diagnostic performance remained similar to the overall external validation cohort (AUC 0.77, 95%CI 0.62-0.92, p<0.001, Supplementary Figure 5).

**AI-CAC diagnostic accuracy according to clinical features**

The AI-CAC diagnostic accuracy measures according to clinical subgroups are reported in Table 3. No interaction of AI-CAC performance in terms of AUCs with clinical subgroups was observed (p-value=ns for all subgroup comparisons, not shown). Overall, AI-CAC specificity tended to be higher in patients of younger age, with less CV risk factors, while the opposite occurred for sensitivity. All AI-CAC accuracy metrics were similar in male and female patients.

Among 254 (49.4%) patients with low/intermediate ESC CV risk, 37.8% had a CAC score >0. AI-CAC was able to detect 84.4% of these patients (p<0.001). Among 260 (50.6%) patients with high/very-high ESC CV risk, 20.4% had a CAC score=0. AI-CAC was able to detect 52.8% of these patients (p<0.001). Among 414 (75.3%) patients with <75 years, 48.8% had a CAC score >0. AI-CAC was able to detect 90.1% of these patients (p<0.001).

**Prognostic value of AI-CAC**

After a median follow-up of 1.5 (IQR 0.8-2.7) years, 17 patients developed an ASCVD event. Both AI-CAC (per AI-CAC category increase, 0/1-99/≥100: HR 2.5, 95%CI 1.2-5.2, p=0.012) and the ESC CV risk grading (HR 2.4, 95%CI 1.3-4.5 per class increase, p=0.004) were associated with 5-yr ASCVD. At bivariate analysis, AI-CAC was associated with borderline significance with 5-year ASCVD (adj-HR 2.2, 95%CI 1.0-4.7, p=0.051), while the ESC CV risk grading was not (adj-HR 1.8, 95%CI 0.6-6.2, p=0.323).

Among 159 patients with AI-CAC=0 a single ASCVD event (non-fatal acute coronary syndrome)

occurred, after 4.3 years from AI-CAC assessment. Patients with AI-CAC>0 had significantly higher Kaplan Meier estimates for the occurrence of ASCVD events as compared to patients with AI-CAC=0 (13.5% vs. 3.4%, log-rank=0.013, Figure 3). A graded increase in ASCVD was observed when stratifying patients in AI-CAC=0, AI-CAC between 1-99 and AI-CAC ≥100 (3.4% vs. 4.9% vs. 23.3%, p=0.025, respectively, Supplementary Figure 6).

Patients with CAC>0 had significantly higher Kaplan Meier estimates for the occurrence of ASCVD events as compared to patients with CAC=0 (23.7% vs. 3.4%, log-rank<0.001, Figure 3).

**Prognostic value of AI-CAC according to clinical features**

Age, CV risk factors and the burden of CV risk factors were associated with ASCVD events (Supplementary Table 4). A graded increase in Kaplan Meier estimate of ASCVD occurrence was observed according to the ESC CV risk grading (0%, 5.2%%, 1.7%, 27.0% from low to very high CV risk, p=0.004, Supplementary Figure 7).

Among patients with low/intermediate ESC CV risk, ASCVD events were similar regardless of AI-CAC (AI-CAC=0 vs. AI-CAC>0: 4.5% vs 3.5%, p=0.249). Conversely, among patients with high/very-high CV risk, those with AI-CAC=0 had no events (as compared to AI-CAC>0: 0.0% vs. 18.4%, p=0.117) (Supplementary Figure 8).

**Model explainability**

Representative maps derived from the Grad-cam method are reported in Supplementary Figure 9. Specifically, each pixel is represented with a color scale reflecting the pixel features importance in the final AI-CAC value output for a given patient. The cardiac silhouette, and particularly the coronary course, and the thoracic aorta seem to inform the AI-CAC model. See Supplementary Figure 9 legend for details."

**DISCUSSION**

Here, we developed a deep-learning based system – the AI-CAC model – to detect subclinical atherosclerosis on plain frontal chest x-ray. The main findings can be summarized as follow:

- The AI-CAC model seems to accurately detect subclinical atherosclerosis on chest x-ray, with sensitivity consistently above 92%.
- The AI-CAC model predicts the 5-year ASCVD risk with high negative predictive value and with similar performance to CAC.
- The prognostic value of AI-CAC seems to be independent from the ESC CV risk grading system.

CAC scoring is the best-established imaging modality to improve CV risk stratification(2). A wealth of evidence highlights a disconnect between the clinical risk profile and the atherosclerotic burden of asymptomatic individuals, translating into modest ASCVD prediction based on the sole clinical risk profile. As CAC score identifies the presence and extent of subclinical coronary atherosclerotic disease rather than its probability (as is the case for clinical risk scores), it presents significant risk reclassification abilities over traditional risk estimators(6,9,10). However, despite its potential to guide public health strategies in primary prevention, its implementation in clinical practice remains limited due to costs, exam accessibility and radiobiological risk concerns.

The use of artificial intelligence technologies in clinical medicine is opening unprecedent opportunities that may be particularly relevant to guide population-based health strategies, where several paradigms of care are expected to change in the near future(14). Specifically, deep-learning based methods are proving their potential to quantify information beyond human perception and their application to simple, inexpensive and widely available exams such as chest x-rays may offer tremendous opportunities across several medical fields(14,21), especially as opportunistic screening strategies.

*Feasibility of CAC detection on chest x-ray.* Our work establishes the feasibility of subclinical atherosclerosis detection on plain frontal chest x-ray based on deep-learning methodologies with diagnostic and prognostic accuracies that merit further prospective testing. A previous work already tested the feasibility of CAC prediction on chest x-ray, showing only moderate accuracy in the internal validation cohort (AUC 0.73) as compared to very good accuracy (AUC 0.90) of the AI-CAC model(22). Similarly, also the prognostic accuracy was moderate, with unsatisfactory negative predictive values for clinical use. Moreover, several limitations comprising inclusion of multiple chest x-ray for a single patient, the pre-processing conversion of high resolution DICOM images to lower resolution JPEG images and the absence of an external validation cohort were reported(22). Inherently to the adopted deep-learning method, the exact features that inform model prediction cannot be inferred with accuracy. However, several hypotheses can be made. First, the coronary calcium information, while generally inaccessible to the human eye on plain chest x-ray, might be contained in specific conformations of pixel density captured by deep-learning methods. Second, several image characteristics that do not necessarily directly reflect coronary calcium might present correlations with CAC, that interpreted with a potent deep-learning methodology able to investigate complex hierarchical representations buried into many hidden layers of information, may produce an accurate output of CAC estimation, as suggested by a recent work evaluating deep-learning CAC prediction by echocardiographic images. The performed explainability analyses, while descriptive in nature, seem to point at the potential role of both hypotheses, as highlighted by the higher information density contributing to model prediction observed along the coronary course, but also in the cardiac silhouette and the aortic root.

*AI-CAC diagnostic performance.* The AI-CAC model demonstrated high sensitivity to rule-out subclinical atherosclerosis, consistently above 92% at both internal and external validation, suggesting its potential role as a gate keeper. As patients with no detectable CAC are at very low risk of ASCVD events (1.5% to 4.9% 10-year ASCVD rate(23)), the benefit of lipid-lowering therapy may be trivial in this subset.(6,9)

Moreover, among patients with low/intermediate ESC CV risk, AI-CAC was able to detect 84.4% of the patients with subclinical atherosclerosis, potentially modifying the treatment approach among those who already developed the ASCVD substratum and that might benefit from more aggressive lipid lowering therapy regardless of clinical risk assessment(6). As specificity was modest (63.9%), the effectiveness and cost-effectiveness of this approach remains to be established, with a specific attention to avoid unnecessary cascade exams and patient psychologic stress. The identification of a grey zone using multiple cut-offs to guide decision making might increase specificity without compromising sensitivity and will be one of the objects of future validation studies.

Of note, while very good diagnostic accuracy was observed at internal validation, the accuracy was somewhat lower at external validation (AUC 0.77, 95%CI 0.67-0.86), driven by lower specificity. While still in the range of fair discrimination, this observation warrants further diagnostic evaluation of the model and, possibly, performance refinement. Importantly, this drop does not appear to be related to model generalizability issues, as the AI-CAC performance was not affected in the subset of images acquired with chest x-ray machine models that were under-represented in the training cohort (12.5%) and constituting about half of the external validation cohort.

*AI-CAC prognostic performance.* The role of AI-CAC to rule-out subclinical atherosclerosis, was further substantiated by its elevated negative predictive value in terms of 5-year ASCVD occurrence. Indeed, a single event (a non-fatal acute coronary syndrome) occurred among patients with an AI-CAC of zero. Importantly, the event occurred after 4.3 years from the initial chest x-ray suggesting a valuable warranty period length following AI-CAC testing.

The AI-CAC prognostic value showed borderline significance after adjusting for the ESC CV risk grading system. This once again suggest the superiority of the atherosclerotic burden over the clinical profile in determining the ASCVD risk(10) supporting its use over clinical risk assessment. Among patients with high/very high ESC CV risk, AI-CAC was able to rule-out subclinical atherosclerosis in one out of two patients with CAC 0. Importantly, these patients had no ASCVD events up to 5-year (vs. 18.4% in patients with high/very high CV risk and CAC >0) despite higher perceived risk based on clinical risk profile. This observation is consistent with large CAC studies showing very low 7-year ASCVD risk (0.3% event/person-year) among patients with ≥3 risk factors with no detectable CAC(24). In this subset, AI-CAC may potentially enhance patient-physician communication, reassure the physician in the clinical decision making of challenging patient categories (i.e. asymptomatic diabetic patients) and direct further diagnostic and therapeutic strategies. Of note, while these patients may still benefit, if clinically indicated, from lipid-lowering therapy in view of their long-term ASCVD risk regardless of the CAC presence, the detection of subclinical atherosclerosis might modify the risk-benefit trade-off of other preventive strategies such as anti-platelet therapy. Indeed, while aspirin allocation guided by the pooled cohort equations may translate in net harm across all ASCVD risk classes, a strategy complemented by CAC evaluation may identify subsets of patients with a risk-benefit trade-off favoring aspirin treatment(25,26).

*AI-CAC clinical use.* Several applications might be conceived to translate the observations of the present analysis to clinical practice, following adequate prospective validation. First, AI-CAC might be used in place of CAC as a cheap, radiation-sparing and widely available diagnostic modality. This may be used in low/intermediate CV risk patients to identify at-risk individuals with subclinical atherosclerosis that may benefit from more aggressive lipid-lowering therapies and in high/very high CV risk patients to down-classify the actual risk. Second, AI-CAC might be used as an opportunistic screening tool among individuals undergoing chest x-ray for any clinical reason. Opportunistic CAC screening of non-gated chest CT followed by clinician and patient notification led to a significant increase in statin prescriptions in a small randomized trial, suggesting the feasibility and utility of this strategy(27). The application of this concept to chest x-ray may tremendously widen the target population with potentially impactful implications on primary prevention public health strategies.

In conclusion, we provide preliminary evidence of the feasibility and diagnostic accuracy of deep-learning based detection of subclinical atherosclerosis on chest x-ray. While the diagnostic and therapeutic implications of a given AI-CAC result are beyond the scope of this feasibility work, our results suggest that, provided adequate testing will confirm the current findings, CAC detection on chest x-ray might ultimately result in clinical utility, including CV risk stratification refinement, appropriate preventive therapies allocation and cost-effectiveness for all the stakeholders (patient, hospitals, insurances, countries).

**Limitations**

The results of the present work should be interpreted in light of several limitations. First, while the sample size of the population is powered for model development and diagnostic accuracy testing, it is limited for the prognostic comparison. Accordingly, its results should be considered exploratory. Second, data on family history, an important CV risk modifier, was not collected. However, family history only marginally improves ASCVD risk prediction beyond conventional risk factors(2). Third, CAC was mostly derived from non-gated CT scan, and due to the limited number of gated-CT scan (<5%), we could not perform sensitivity analysis. While gated-CT scan represents the gold standard for CAC evaluation, a wealth of evidence supports the high agreement of CAC measurement from non-gated CT scan with that of gated CT scan (in the Multi-Ethnic Study of Atherosclerosis agreement substudy: r statistic 0.95; sensitivity 96.8% and specificity 92.5% for no CAC vs. CAC >0)(28) along with its robust prognostic value(29). Moreover, slice thickness has been previously identified as the main source of discordance between standard and gated chest CT scan(30); and we only included chest CT scan with slices ≤2.5 mm, that is below conventional 3 mm gated CT scan for CAC scoring.

Fourth, the cohort adopted for model development included patients with available chest X-ray and CT-scan indicated for any clinical reason. While careful clinical data collection was carried to exclude known ASCVD, and the cohort demographics are in line with the target population, the patient characteristics may still partly diverge from those of primary prevention individuals with an indication for CAC scoring. Of note, Kaplan Meier curves denote a somewhat higher ASCVD incidence in the first year following chest x-ray assessment, that would not be expected in a general primary prevention population. This may reflect a bias due to either incidental findings on chest CT scan (i.e. coronary calcifications) prompting further diagnostic work-up and eventually coronary revascularization, or due to symptomatic patients (i.e. dyspnea) undergoing chest x-ray/CT scan as part of diagnostic work-up and eventually undergoing coronary revascularization. This observation doesn't affect the primary analytical purpose of diagnostic accuracy, and it is also unlikely to have significantly affected the prognostic comparison, especially in terms of the demonstrated independency of the AI-CAC prognostic value from the clinical risk profile. Moreover due to the heterogeneous nature of chest x-ray indications, we believe that it is unlikely that any systematic confounding might have influenced model development. Finally, the external validation cohort comprises patients from the same institution collected during a subsequent historical period, has a modest sample size, and – despite still in the range of good accuracy – the diagnostic performance appears numerically lower as compared to the internal validation cohort. Thus, a prospective, multi-center real-world study assessing the accuracy of the AI-CAC score in the target primary prevention population is warranted. The AI-CAC score Prospective Validation Study, assessing the diagnostic accuracy of the AI-CAC score in a multi-center cohort of primary prevention patients with a clinical indication for CAC testing, has been funded, is about to start patient enrollment, and will establish the score generalizability and potential clinical value.

**CONCLUSIONS**

The deep-learning based AI-CAC model seems to accurately detect subclinical atherosclerosis on chest x-ray and to predict ASCVD events independently from the CV risk profile. Adoption of the AI-CAC model to refine CV risk stratification or as an opportunistic screening tool requires prospective evaluation.

**ACKNOWLEDGEMENTS.** None.

**FUNDING.** None.

**CONFLICT OF INTERESTS.** G.G., F.I., A.P., O.D.F., W.G.M., J.M.H., M.I., P.F., A.F., M.G., and F.D.A. have submitted a patent application, pending, covering the AI-CAC model. The other authors declare no conflict of interests.

**AUTHORS' CONTRIBUTIONS.** GG, FDA and GMDF designed the study. DT, MV, AS, PG and AB collected the data. AP, CAB, AF and MG took part in the development of the model. GG and AP performed the statistical analysis. GG drafted the manuscript. All authors contributed significantly to the writing and critical review of the manuscript and approved the final draft.

**DATA AVAILABILITY STATEMENT.** The AI-CAC development dataset is not publicly available at this moment, and all research or research- related activities that involve an external party might require a written research agreement to define obligations and manage the risks.

TABLES

**Table 1. Demographics, CV risk profile and CAC score data overall and stratified by no AI-CAC vs. AI-CAC >0.**

Abbreviations: CAC: coronary artery calcium; CKD: chronic kidney disease; CV: cardiovascular; eGFR: estimated glomerular filtration rate.

|  | POOLED STUDY COHORTS | | | |
|---|---|---|---|---|
|  | Overall (n=550) | AI-CAC =0 (n=159) | AI-CAC >0 (n=391) | p-value |
| **Demographics** | | | | |
| Age (yrs) | 63 (51-74) | 48 (35-60) | 69 (59-77) | <0.001 |
| Male sex (%) | 58.4 (321) | 47.8 (76) | 62.7 (245) | 0.001 |
| **CV risk profile** | | | | |
| Smoke (%) | 48.1 (247) | 33.6 (50) | 54.0 (197) | <0.001 |
| Diabetes (%) | 21.4 (110) | 11.4 (17) | 25.5 (93) | <0.001 |
|    Diabetes duration ≥10 years (%) | 15.6 (80) (59) | 5.3 (8) | 19.7 (72) | <0.001 |
|    Diabetes with organ damage (%) | 11.7 (60) | 2 (3) | 15.6 (57) | <0.001 |
| Hypertension (%) | 46.1 (237) | 22.8 (34) | 55.6 (203) | <0.001 |
|    Severe hypertension (%) | 1.9 (10) | 0.0 (0) | 2.7 (10) | <0.001 |
| Dyslipidemia (%) | 24.3 (125) | 12.1 (18) | 29.3 (107) | <0.001 |
|    Severe dyslipidemia (%) | 1.4 (7) | 0 (0) | 1.9 (7) | 0.090 |
|    Lipid-lowering therapy (%) | 21.4 (110) | 9.4 (14) | 26.3 (96) | <0.001 |
| Severe CKD (GFR <30 ml/min/mq) | 7.0 (36) | 3.4 (5) | 8.5 (31) | 0.025 |
| Subclinical atherosclerosis (%) | 10.1 (52) | 1.3 (2) | 13.7 (50) | <0.001 |
| **Number of CV risk factors** | | | | |
|    No | 27.2 (149) | 47.2 (75) | 18.9 (74) |  |
|    1-2 | 57.5 (316) | 49.1 (78) | 60.9 (238) |  |
|    >2 | 15.5 (85) | 3.7 (6) | 21.1 (78) | <0.001 |
| **ESC CV risk category** | | | | |
|    Low (%) | 4.1 (21) | 12.1 (18) | 0.8 (3) |  |
|    Intermediate (%) | 45.3 (233) | 66.4 (99) | 36.7 (134) |  |
|    High (%) | 26.3 (135) | 14.1 (21) | 31.2 (114) |  |
|    Very high (%) | 24.3 (125) | 7.4 (11) | 31.2 (114) | <0.001 |
| **CAC assessment** | | | | |
| CAC score | 54 (0-894) | 0 (0-0) | 336 (11-1433) | <0.001 |
| AI-CAC score | 35 (0-388) | 0 (0-0) | 218 (17-670) | <0.001 |

**Table 2. Sensitivity, specificity, positive and negative predictive values for an AI-CAC value >0 to detect a CAC >0 in the derivation, internal validation and external validation cohorts.**

Abbreviations: AUC: area under the curve; NPV: negative predictive value; PPV: positive predictive value.

|  | AUC (95%CI) | Sensitivity | Specificity | NPV | PPV |
|---|---|---|---|---|---|
| Derivation cohort (n=460) | 0.88 (0.85-0.91) | 92.8% | 63.9% | 85.4% | 79.6% |
| Internal validation cohort (n=92) | 0.90 (0.84–0.97) | 94.6% | 69.4% | 82.3% | 82.8% |
| External validation cohort (n=90) | 0.77 (0.67–0.86) | 97.8% | 45.5% | 95.2% | 65.2% |

**Table 3. Sensitivity, specificity, positive and negative predictive values for an AI-CAC value >0 to detect a CAC >0 in the overall cohort, stratified by clinical subgroups.**

Abbreviations as in Tables 1 and 2.

|  | AUC (95%CI) | Sensitivity | Specificity | NPV | PPV |
|---|---|---|---|---|---|
| **Age** | | | | | |
| Age ≥75 years | 0.78 (0.65-0.91) | 98.9% | 28.6% | 75.0% | 92.0% |
| Age <75 years | 0.84 (0.80-0.88) | 87.2 % | 63.6 % | 87.3% | 63.2% |
| **ESC CV risk category** | | | | | |
| Low to intermediate CV risk | 0.81 (0.75-0.86) | 84.4% | 63.9% | 87.1% | 58.7% |
| High to very high CV risk | 0.84 (0.77-0.90) | 98.1% | 52.8% | 87.5% | 89.0% |
| **Sex** | | | | | |
| Female sex | 0.87 (0.83-0.91) | 90.2% | 60.7% | 86.6% | 68.7% |
| Male sex | 0.84 (0.78-0.89) | 95.3 % | 60.0 % | 86.8% | 82.0% |
|  | | | | | |
| Diabetes | 0.87 (0.79-0.95) | 97.6% | 60.0% | 88.2% | 89.2% |
| No diabetes | 0.85 (0.81-0.89) | 92.2% | 71.3% | 87.0% | 73.6% |
| **CKD** | | | | | |
| Severe CKD (eGFR <30 | 0.87 (0.74-0.99) | 96.4% | 50.0% | 80.0% | 87.1% |
| No severe CKD (eGFR <30 | 0.86 (0.83-0.89) | 93.5% | 61.6% | 87.4% | 76.7% |
| **Number of CV risk factors** | | | | | |
| No CV risk factors | 0.83 (0.76-0.90) | 86.3% | 68.4% | 90.5% | 58.7% |
| 1-2 CV risk factors | 0.84 (0.80-0.89) | 92.9% | 54.2% | 82.1% | 77.3% |
| ≥2 CV risk factors | 0.89 (0.78-0.99) | 100.0% | 54.5% | 100% | 93.7% |

**FIGURE LEGENDS**

**Figure 1. Workflow of the image preprocessing pipeline.** Starting from the original DICOM file, the final input tensor was obtained following all the represented preprocessing stages. The stages are detailed in the supplementary methods of the supplementary appendix.

**Figure 2. Area under the curve (AUC) for the AI-CAC model to predict a CAC >0 in the internal validation and external validation cohorts.**
Abbreviations: AUC: area under the curve; CI: Confidence interval.

**Figure 3. Kaplan Meier estimates for ASCVD events among patients with AI-CAC >0 vs AI-CAC =0 (left), and among patients with CAC >0 vs CAC =0 (right).**

**Figure 4. Representative importance maps derived from the integrated gradient method.**
In these importance maps each pixel is represented with a gray scale degree reflecting the pixel features importance in the final AI-CAC value output for a given patient. Four representative maps are reported, coupled with an informative slice of the relating chest CT scan. The darker pixels in each map represent the anatomical structures in whom the model relied for that specific patient to elaborate the AI-CAC value. Overall, several anatomical structures including the cardiac silhouette (and particularly the coronary course) and the thoracic aorta seem to inform the AI-CAC model.

**Graphical abstract. Feasibility and performance of deep-learning based detection of subclinical atherosclerosis on chest x-ray.** Left panel: a deep-learning algorithm to predict CAC score on chest radiograms (the AI-CAC model) was retrospectively developed on 460 chest x-rays of primary prevention patients with available paired chest x-ray and chest computed tomography (CT), using the CAC score calculated on chest CT as ground truth. Right panel: the AI-CAC model accurately detects subclinical atherosclerosis on chest x-ray and predicts atherosclerotic cardiovascular disease events with elevated negative predictive value. Abbreviations: AI: artificial

intelligence; ASCVD: atherosclerotic cardiovascular disease; AUC: area under the curve; CAC: coronary artery calcium; CI: Confidence Interval.

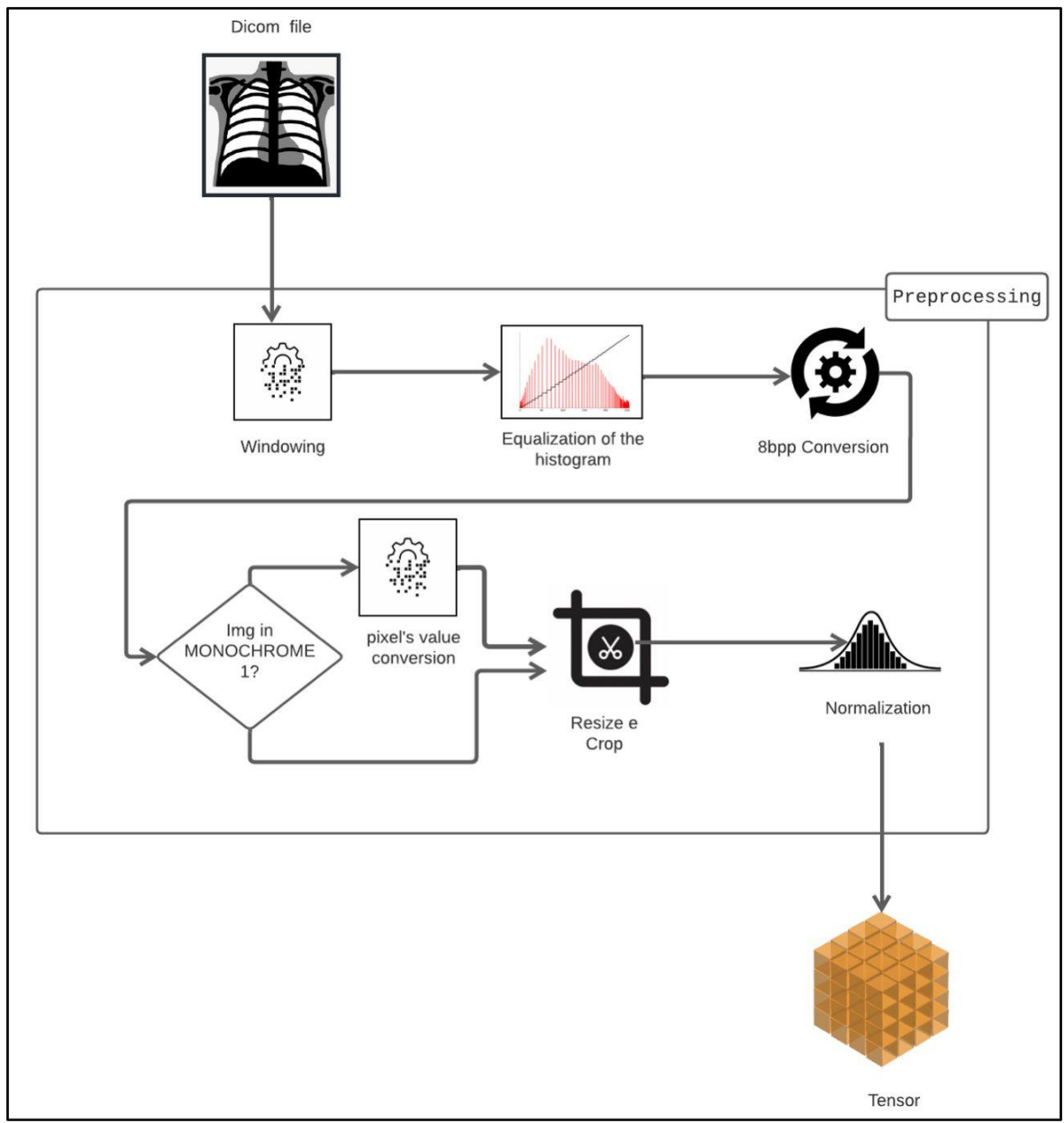

***Figure 1 :*** *Workflow of the image preprocessing pipeline. Starting from the original DICOM file, the final input tensor was obtained following all the represented preprocessing stages. The stages are detailed in the supplementary methods of the supplementary appendix.*

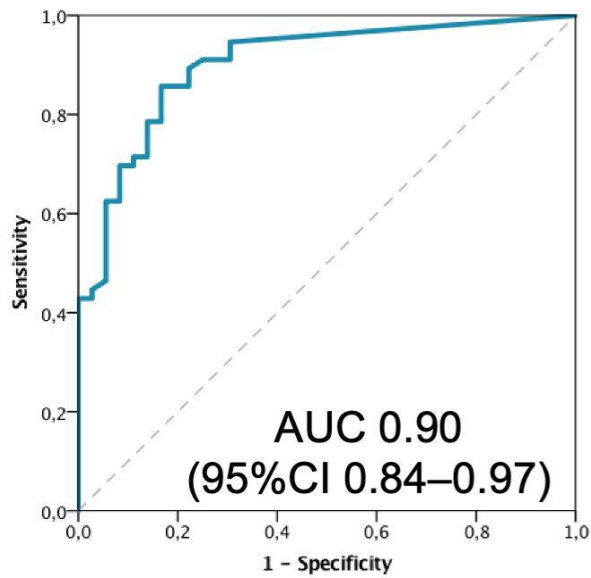 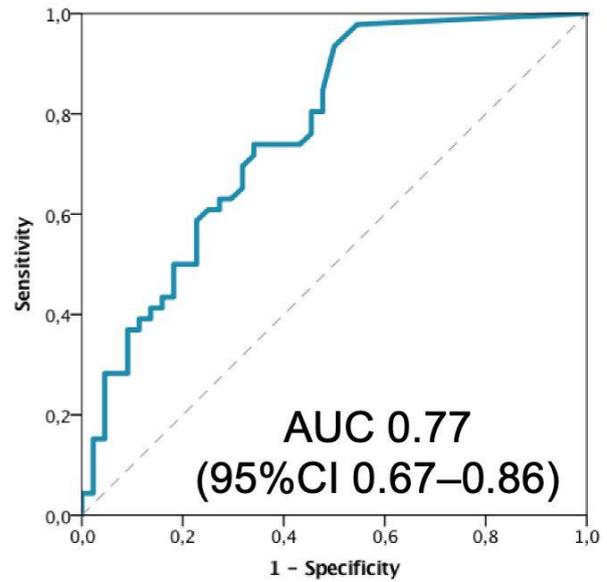

**Figure 2** : Area under the curve (AUC) for the AI-CAC model to predict a CAC >0 in the internal validation and external validation cohorts. Abbreviations: AUC: area under the curve; CI: Confidence interval.

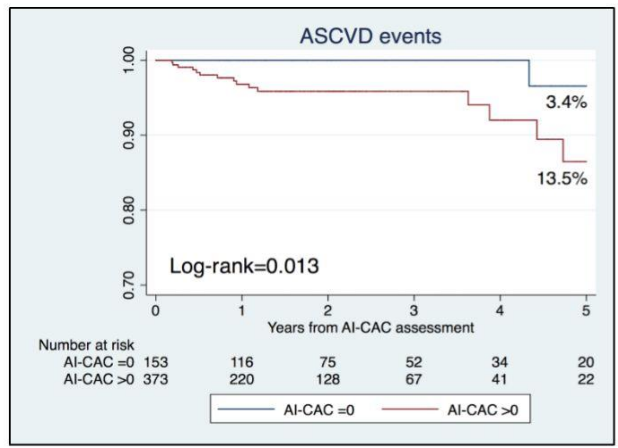 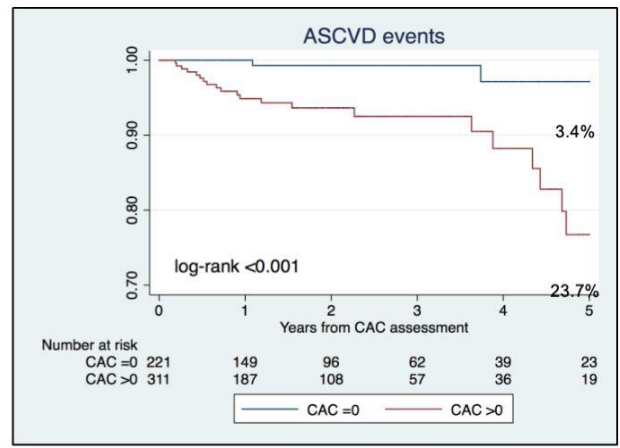

**Figure 3** : Kaplan Meier estimates for ASCVD events among patients with AI-CAC >0 vs AI-CAC =0 (left), and among patients with CAC >0 vs CAC =0 (right).

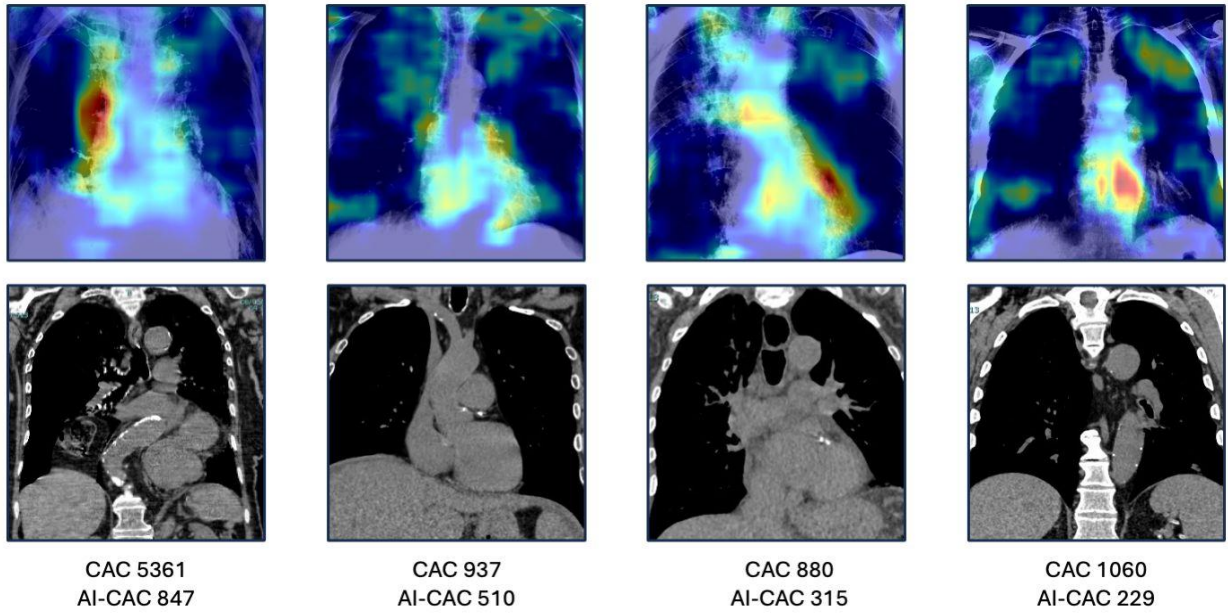

**Figure 4.** Representative importance maps derived from the integrated gradient method. In these importance maps each pixel is represented with a gray scale degree reflecting the pixel features importance in the final AI-CAC value output for a given patient. Four representative maps are reported, coupled with an informative slice of the relating chest CT scan. The darker pixels in each map represent the anatomical structures in whom the model relied for that specific patient to elaborate the AI-CAC value. Overall, several anatomical structures including the cardiac silhouette (and particularly the coronary course) and the thoracic aorta seem to inform the AI-CAC model.

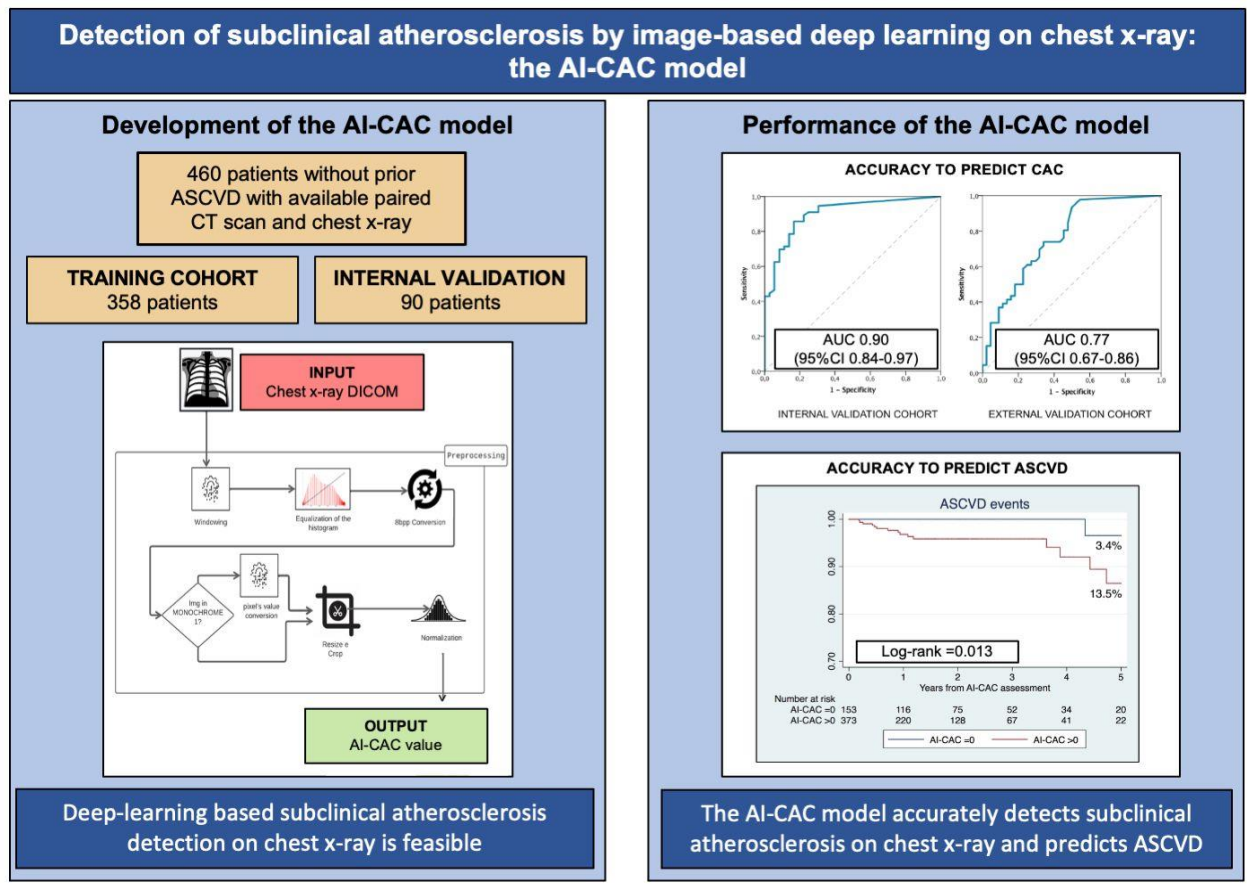

*Figure 5:* Graphical abstract. Feasibility and performance of deep-learning based detection of subclinical atherosclerosis on chest x-ray. Left panel: a deep-learning algorithm to predict CAC score on chest radiograms (the AI-CAC model) was retrospectively developed on 460 chest x-rays of primary prevention patients with available paired chest x-ray and chest computed tomography (CT), using the CAC score calculated on chest CT as ground truth. Right panel: the AI-CAC model accurately detects subclinical atherosclerosis on chest x-ray and predicts atherosclerotic cardiovascular disease events with elevated negative predictive value. Abbreviations: AI: artificial intelligence; ASCVD: atherosclerotic cardiovascular disease; AUC: area under the curve; CAC: coronary artery calcium; CI: Confidence Interval.

# SUPPLEMENTARY APPENDIX

# Detection of subclinical atherosclerosis by image-based deep learning on chest x-ray

# SUPPLEMENTARY METHODS

In the following section, the methods followed to develop the AI-CAC model are described, detailing the data preparation process, the architecture of the neural network tasked with prediction, and the relative training procedure.

## 1.1 Image preprocessing

Radiographic images were preprocessed before they could be fed into a neural network. As radiographic images are encoded as single-channel grayscale images, for each image, the following preprocessing steps were taken:

- **Windowing**: the grayscale image was processed through a LUT operation in order to highlight particular structures. In particular, the metadata called *WindowWidth* and *Windowcenter* (stored as metadata in the dicom file), which regulate the center and the length of the window respectively, were exploited to obtain the opportune range of representation. After this step, pixels were distributed in the range from *Windowcenter – (WindowWidth)/2* to *Windowcenter + (WindowWidth)/2*
- **Contrast enhancement** by equalization of the histogram.
- **Resizing of the image** to a dimension of 1248x1248 pixels, exploiting bilinear interpolation, and a **central cropping**, obtaining thus a final image of 1024X1024 pixels: these two final steps were necessary to make all images the same size and to be able to feed them to the model.
- **Standardization** of the data exploiting the mean $\mu$ and the standard deviation $\sigma$ computed over a training set. After $\mu$ and $\sigma$ were calculated, the following formula was applied to all pixels of the image:

$$(1) \quad \mu = \frac{1}{W \cdot H} \sum_{i,j} x_{ij} \; \forall \, i = 1....N, \, j = 1....M$$

$$(2) \quad \sigma = \frac{1}{W \cdot H} \sum_{i,j} (x_{ij} - \mu)^2 \; \forall \, i = 1....N, \, j = 1....M$$

$$(3) \quad x = (x - \mu)/\sigma$$

where W and H represented respectively the width and height of the image.

A scheme of this process is represented in **Supplementary Figure 1**.

## 1.2 Neural network architecture

The architecture of the model to predict the calcium score consisted of a deep convolutional feature extractor followed by one prediction head. Concerning the features extractor, we followed a similar approach to [4] relying on a DenseNet121 [2] backbone. The DenseNet architecture is composed of 120 convolutional layers and 1 final fully connected layer, which has been removed as it is not useful for this task: after an initial 7x7 convolution layer and a 2x2 max pooling layer, the net is organized in 4 different dense blocks, composed mainly by 1x1 and 3x3 convolution layers, and 3 transition layers, each of them composed by 1x1 convolution layer and 2x2 average pooling layer. DenseNet builds upon and improves over ResNet [6], in that skip connections are leveraged so that each layer receives in inputs all the feature maps produced by all the previous layers: in fact the role of dense blocks and transitions layers is to adapt the dimensions of the different feature maps to be able to concatenate them. As a result, the network can be thinner and compact with fewer channels with respect to ResNet, thus having both high computational and memory efficiency. The output of the DenseNet feature extractor is a vector of 1024 features maps sized 32x32; an average pooling layer then flattens the features map, producing in output a vector of 1024 features that is the representation of the input image.

Notice that we experimented with different backbones like ResNet [6] or EfficientNet [7], however they yield poorer results. Concerning the prediction head, it consists of a fully connected hidden layer with 64 neurons with ReLu [5] activation function and one output layer. Since the network is expected to yield a real-valued number representing the calcium score, the output layer includes only one neuron with linear activation. Notice that we experimented with different prediction head topologies, activation functions, and techniques like dropout [8], however these experiments never yielded any benefit despite the increased complexity.

**1.3 Ground truth preparation**

In order to address the regression task, as the range of the calcium score is quite broad and this could affect the stability of the algorithm during the learning process, we took the following steps:

- We clipped the calcium score to 2000.
- We transformed the original domain in logarithmic scale, i.e. we took the logarithm of the calcium score for each sample of the dataset with the following formula:

$$y' = log(y)$$

where $y$ is the original value and $y'$ is the new one.

- We normalized the labels over $\mu'$ and $\sigma'$, which are the mean and the standard deviation computed in the new logarithmic domain

This normalized domain has allowed the model to generalize well on the data.

We also transformed the threshold in the logarithmic domain with the following formula:

$$th' = (log(th + \epsilon) - \mu')/\sigma' \quad (4)$$

Next, label $y'$ and the predicted value $\underline{y'}$ were compared exploiting $th'$: if both $y'$ and $\underline{y'}$ are above or below this threshold, the sample is correctly classified. $\epsilon = 0.00001$ is necessary to avoid logarithm if 0 if the threshold is set to be 0.

**1.4 Training procedure**

As a first step, we pretrained the feature extractor on the Chexpert dataset [3] following the same supervised procedure as in [4]. The choice to pretrain on Chexpert is due to the fact that the DenseNet121 includes more than 7 million parameters and it was originally designed to be trained over the much larger datasets, such as ImageNet[11] that encompasses over 1 million images. To the best of our knowledge, Chexpert is one of the largest chest X-rays datasets, with a total of 224,316 images from 65,240 patients.

Next, concerning the actual supervised training, the dataset (derivation cohort) was randomly split into 80% training data, and the remaining 20% was used for testing (internal validation).

The Mean absolute error (MAE) was minimized as:

(6) MAE = $\frac{1}{N} \sum_{i=1}^{N} | y_i - \underline{y}_i |$.

In the above formulas, N, $y_i$, and $\underline{y}_i$ represents the total number of samples in a batch, the ground truth for the i-th sample and $\underline{y}_i$ the network prediction respectively. The training hyperparameters are shown in Supplementary Table 1. The adopted optimization algorithms are reported in the following table.

Furthermore, we found that best results could be achieved when training the prediction head and the last block of the DenseNet121 feature extractor only, while keeping the rest of the learnable parameters frozen. That is, training the remaining layers of the feature extractor would bring no additional gains despite increased training time. We hypothesize that the ChexPert pre-training was successful in learning lower-level feature detectors that are also suitable for our calcium score application.

## SUPPLEMENTARY RESULTS

The AI-CAC model was cross-validated over 5 different folds [9]. Results are presented according to different performance metrics.

Concerning the regression task, resorting to the threshold defined in sec. 1.3, Supplementary Table 2 shows that the average balanced accuracy is around 0.78, with a sensitivity of 0.81. For this task, instead of the AUC we provide the Regression AUC (RAUC) [7] values in place of AUC.

It could be interesting to assess whether the calcium Regression model is able to output plausible values of the calcium. **Supplementary Figure 2** shows the errors of the regressor for each sample of the test of the second fold. On the x axis are reported the validation samples in increasing order of ground truth calcium score. On the y axis are reported the values predicted by the network in the logarithmic scale and normalized with $\mu\_log$ and $\sigma\_log$. Vertical lines represent the error for each sample, while the red horizontal line represents the normalized threshold between the two classes we imposed. The error of the regressor is not negligible, despite the corresponding classification accuracy is not significantly affected.

# SUPPLEMENTARY TABLES

**Supplementary Table 1. Main hyperparameters used for training the calcium score predicting network, depending on the task.**

| Hyperparameter | Continuous regression |
|---|---|
| Epochs | 80 |
| Optimizer | Stochastic gradient descent |
| Loss | MAE |
| Learning rate | 0.0003 |
| Weight decay | 0.0001 |
| Batch size | 4 |
| Threshold | 0 |

**Supplementary Table 2. Calcium score prediction results for the continuous regression task on the validation samples**

| Fold | Accuracy | Bal. accuracy | Sensitivity | Specificity | RAUC |
|---|---|---|---|---|---|
| 1 | 0.76 | 0.75 | 0.80 | 0.69 | 0.65 |
| 2 | 0.83 | 0.82 | 0.83 | 0.82 | 0.76 |
| 3 | 0.75 | 0.73 | 0.80 | 0.79 | 0.60 |
| 4 | 0.82 | 0.82 | 0.81 | 0.81 | 0.67 |
| 5 | 0.80 | 0.78 | 0.84 | 0.72 | 0.78 |
| Mean | **0.79** | **0.78** | **0.81** | **0.76** | **0.69** |

**Supplementary Table 3.** Demographics, CV risk profile, and CAC score data overall and stratified by no AI-CAC vs. AI-CAC >0, in the derivation and in the external validation cohorts.

|  | DERIVATION COHORT | | | | EXTERNAL VALIDATION COHORT | | | | p-value derivation vs. external validation cohorts |
|---|---|---|---|---|---|---|---|---|---|
|  | Overall (n=460) | AI-CAC =0 (n=137) | AI-CAC >0 (n=323) | p-value | Overall (n=90) | AI-CAC =0 (n=21) | AI-CAC >0 (n=69) | p-value | |
| **Demographics** | | | | | | | | | |
| Age (yrs) | 64 (52-75) | 50 (37-62) | 69 (60-78) | <0.001 | 59 (45-72) | 33 (20-46) | 64 (55-76) | <0.001 | 0.017 |
| Male sex (%) | 57.8 (266) | 44.5 (61) | 63.5 (205) | <0.001 | 61.1 (55) | 68.2 (15) | 58.8 (40) | 0.301 | 0.324 |
| **CV risk profile** | | | | | | | | | |
| Smoke (%) | 48.5 (207) | 32.3 (41) | 55.3 (166) | <0.001 | 46.0 (40) | 40.9 (9) | 47.7 (31) | 0.382 | 0.380 |
| Diabetes (%) | 22.0 (94) | 11.8 (15) | 26.3 (79) | <0.001 | 18.4 (16) | 9.1 (2) | 21.5 (14) | 0.163 | 0.276 |
| Diabetes duration ≥10 years (%) | 17 (73) | 6.3 (8) | 21.7 (65) | <0.001 | 8.0 (7) | 10.8 /7) | 10.8 (7) | 0.134 | 0.056 |
| Diabetes with organ damage (%) | 11.9 (51) | 2.4 (3) | 16.0 (48) | <0.001 | 10.3 (9) | 0.0 (0) | 13.8 (9) | 0.062 | 0.417 |
| Hypertension (%) | 48.2 (206) | 26.0 (33) | 57.7 (173) | <0.001 | 35.6 (31) | 4.5 (1) | 46.2 (30) | <0.001 | 0.020 |
| Severe hypertension (%) | 1.9 (8) | 0.0 (0) | 2.7 (8) | <0.001 | 2.3 (2) | 0.0 (0) | 3.1 (2) | 0.001 | 0.467 |
| Dyslipidemia (%) | 24.8 (106) | 14.2 (18) | 29.3 (88) | <0.001 | 21.8 (19) | 0.0 (0) | 29.2 (19) | 0.002 | 0.329 |
| Severe dyslipidemia (%) | 1.6 (7) | 0.0 (0) | 2.3 (7) | 0.083 | 0 (0) | 0 (0) | 0 (0) | 1.000 | 0.271 |
| Lipid-lowering therapy (%) | 21.8 (93) | 11.0 (14) | 26.3 (79) | <0.001 | 19.5 (17) | 0.0 (0) | 26.2 (17) | 0.004 | 0.381 |
| Severe CKD (GFR <30 ml/min/mq) (%) | 7.5 (32) | 3.1 (4) | 9.3 (28) | 0.017 | 4.6 (4) | 4.5 (1) | 4.6 (3) | 0.736 | 0.238 |
| Subclinical atherosclerosis (%) | 11.2 (48) | 1.6 (2) | 15.3 (46) | <0.001 | 4.6 (4) | 0.0 (0) | 6.2 (4) | 0.304 | 0.039 |
| **Number of CV risk factors** | | | | | | | | | |
| No | 26.5 (122) | 46.0 (63) | 18.3 (59) | <0.001 | 30.0 (27) | 54.5 (12) | 22.1 (15) | 0.001 | |
| 1-2 | 57.6 (265) | 49.7 (68) | 61.0 (197) | <0.001 | 56.7 (51) | 45.0 (10) | 60.3 (41) | 0.001 | 0.417 |
| >2 | 15.9 (73) | 4.3 (6) | 20.8 (67) | <0.001 | 13.3 (12) | 0.0 (0) | 17.7 (12) | 0.001 | |
| **ESC CV risk category** | | | | | | | | | |
| Low (%) | 4.4 (19) | 13.4 (17) | 0.7 (2) | <0.001 | 2.3 (2) | 4.5 (1) | 1.5 (1) | 0.002 | |
| Intermediate (%) | 42.2 (180) | 62.2 (79) | 33.7 (101) | <0.001 | 60.9 (53) | 90.9 (20) | 50.8 (33) | 0.002 | 0.019 |
| High (%) | 27.4 (117) | 16.5 (21) | 32.0 (96) | <0.001 | 20.7 (18) | 0.0 (0) | 27.7 (18) | 0.002 | |
| Very high (%) | 26.0 (111) | 7.9 (10) | 36.7 (101) | <0.001 | 16.1 (14) | 4.5 (1) | 20.0 (13) | 0.002 | |
| **CAC assessment** | | | | | | | | | |
| CAC score | 76 (0-963) | 0 (0-0) | 428 (22-1576) | <0.001 | 48 (1-337) | 0 (0-0) | 85 (0-931) | <0.001 | 0.049 |
| AI-CAC score | 29 (0-404) | 0 (0-0) | 233 (17-750) | <0.001 | 4 (0-447) | 0 (0-0) | 168 (20-530) | <0.001 | 0.140 |

Abbreviations: CAC: coronary artery calcium; CKD: chronic kidney disease; CV: cardiovascular; eGFR: estimated glomerular filtration rate.

**Supplementary Table 4.** Univariate predictors of ASCVD in the overall cohort.

|  | HR (95%CI) | p-value |
|---|---|---|
| Male sex | 0.55 (0.23-1.31) | 0.177 |
| Age ≥75 years | 2.41 (1.04-5.58) | 0.040 |
| Dyslipidaemia | 3.49 (1.53-7.97) | **0.003** |
| Smoke | 2.78 (1.14-6.78) | **0.024** |
| Hypertension | 2.08 (0.88-4.92) | 0.094 |
| Severe CKD | 2.99 (1.02-8.81) | **0.046** |
| Diabetes | 2.69 (1.14-6.17) | **0.024** |
| ESC CV risk class (per class increase) | 2.69 (1.58-4.56) | **<0.001** |
| N. of CV RF (per 2RF increase, reference 0 RF) | 3.21 (1.63-6.30) | **0.001** |

Abbreviations: CI: confidence intervals; CKD: chronic kidney disease; CV: cardiovascular; HR: hazard ratio; RF: risk factor.

# SUPPLEMENTARY FIGURES

**Supplementary Figure 1. Workflow of the image preprocessing pipeline.** Starting from the original DICOM file, we obtain the final input tensor following all the preprocessing stages presented in Sec. 1.1.

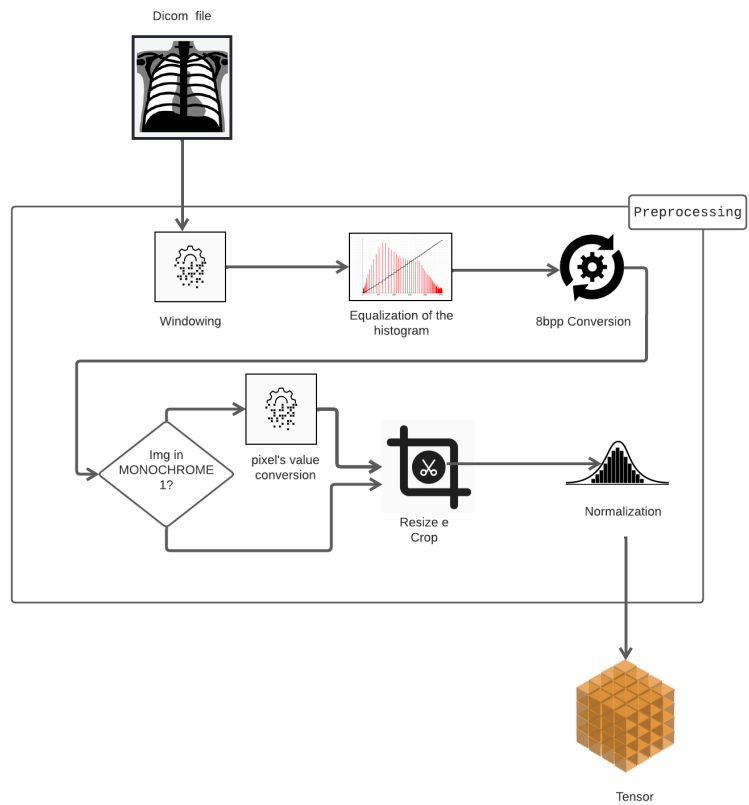

**Supplementary Figure 2. Calibration plots.** Top panel: a calibration plot depicting the predicted (AI-CAC score) and the observed (real CAC score) CAC value in the internal (left) and external (right) validation cohorts is reported. Bottom panel: histograms reporting the predicted (AI-CAC score) and the observed (real CAC score) by real CAC score strata.

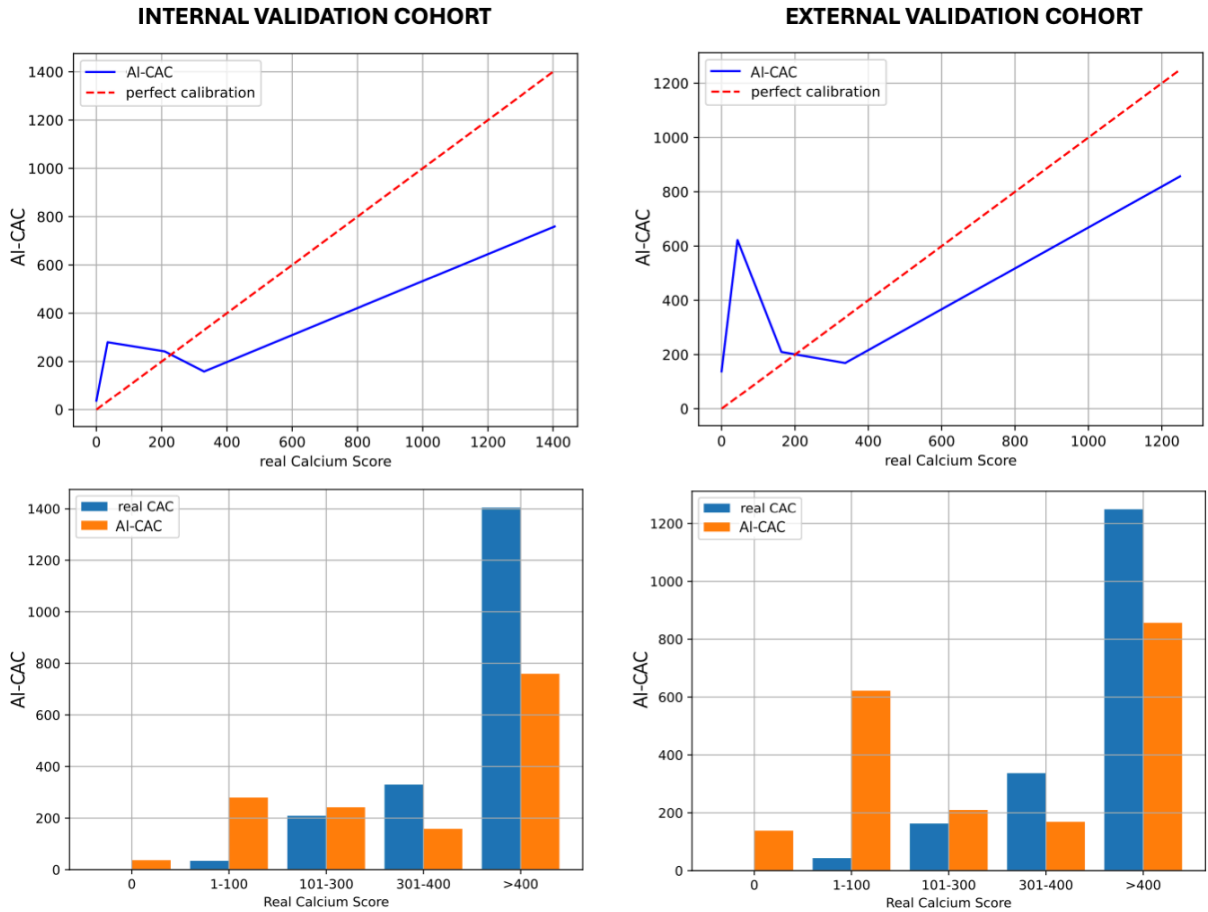

**Supplementary Figure 3. Precision-Recall curves for the AI-CAC model to predict a CAC >0 in the internal validation and external validation cohorts.**

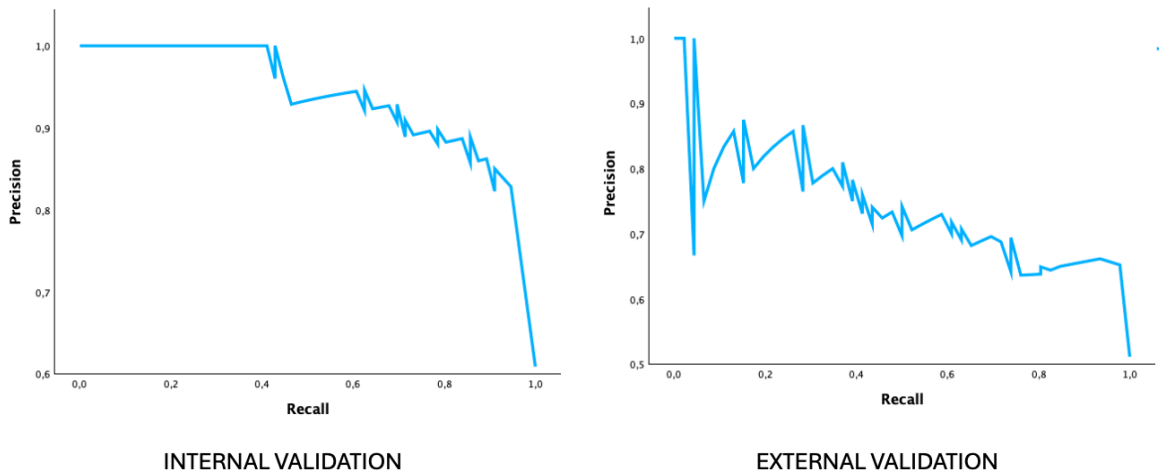

INTERNAL VALIDATION                    EXTERNAL VALIDATION

**Supplementary Figure 4. Distribution of x-ray machine models adopted for image acquisition in the training, internal validation and external validation cohorts.**

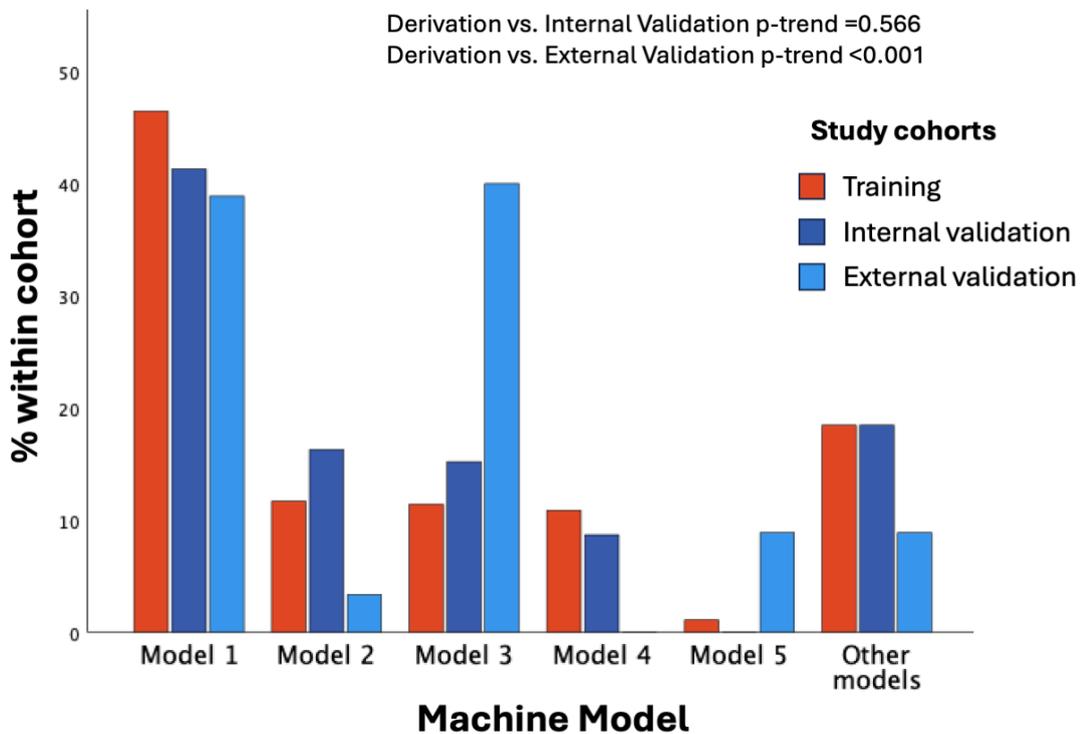

**Supplementary Figure 5.** Area under the curve (AUC) for the AI-CAC model to predict a CAC >0 in the external validation subset of chest x-rays acquired with renovated machine models, under-represented in the derivation cohort.

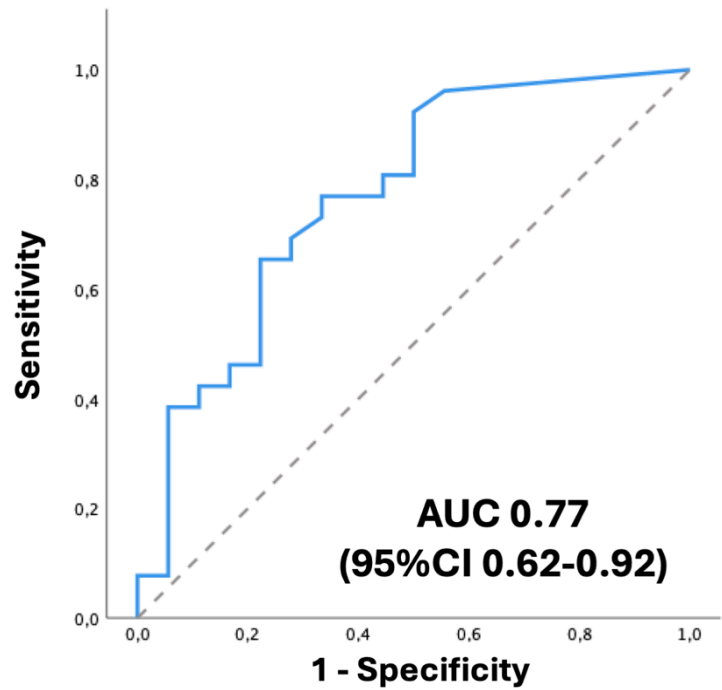

Abbreviations: AUC: area under the curve; CI: Confidence interval.

**Supplementary Figure 6. Kaplan Meier estimates for ASCVD events stratified by AI-CAC categories.**

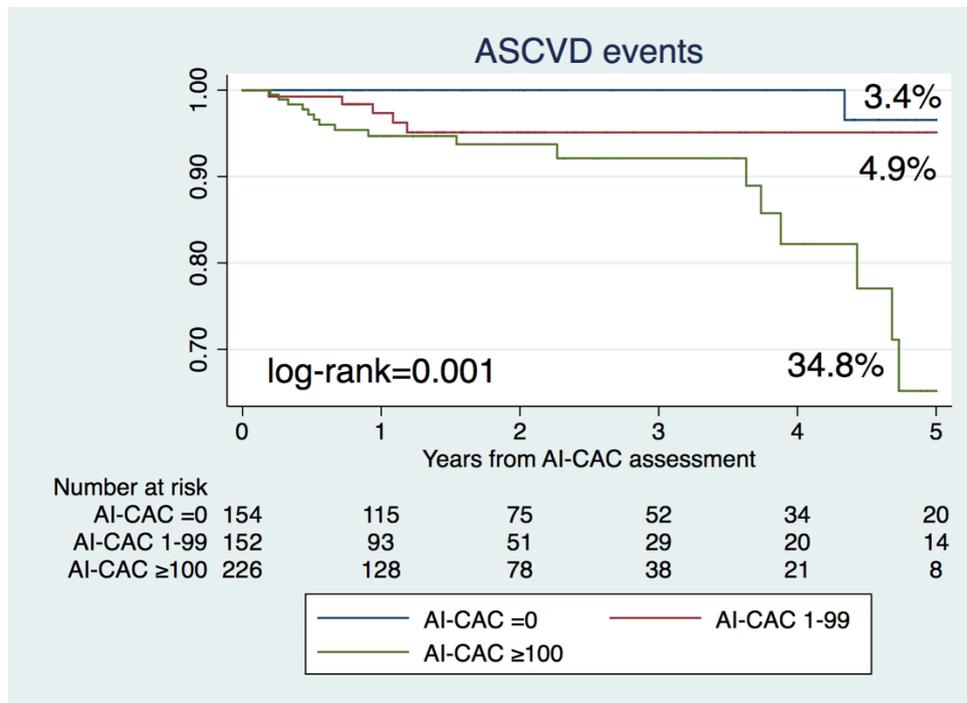

**Supplementary Figure 7. Kaplan Meier estimates for ASCVD events stratified by ESC CV risk categories**

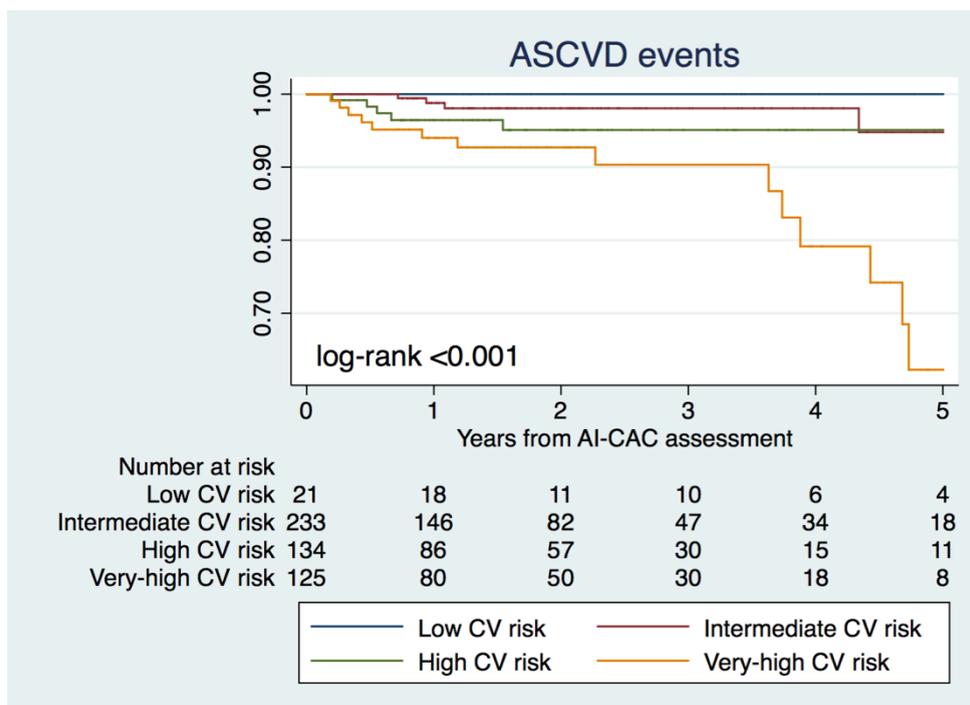

**Supplementary Figure 8. Kaplan Meier estimates for ASCVD events stratified by concordant and discordant AI-CAC and ESC CV risk categories.**

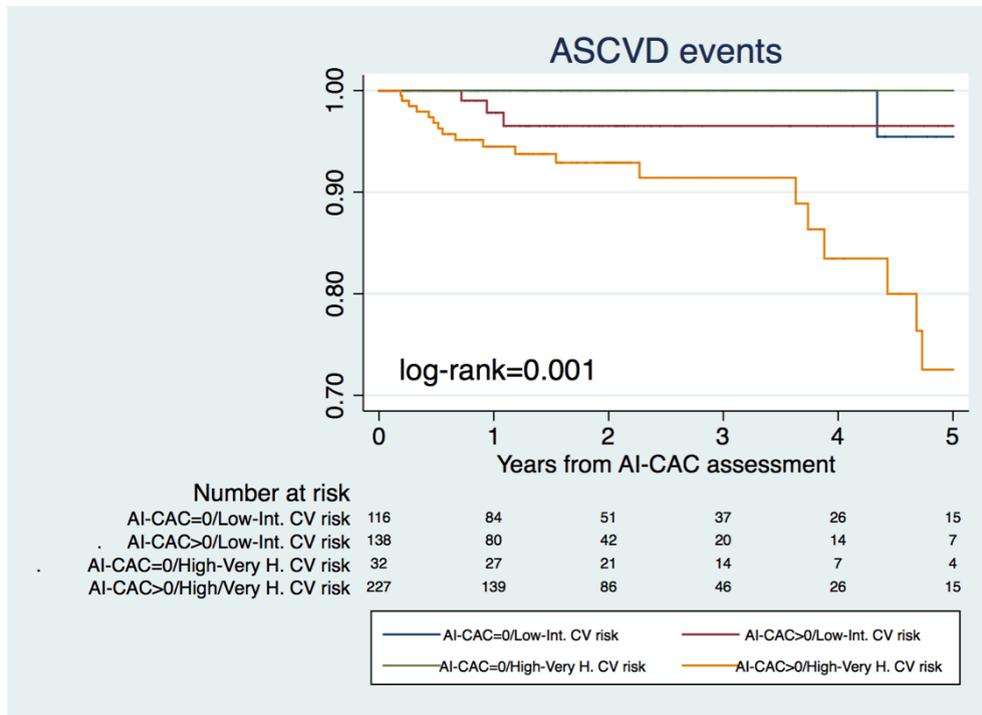

**Supplementary Figure 9. Representative maps derived from the GradCam method.**

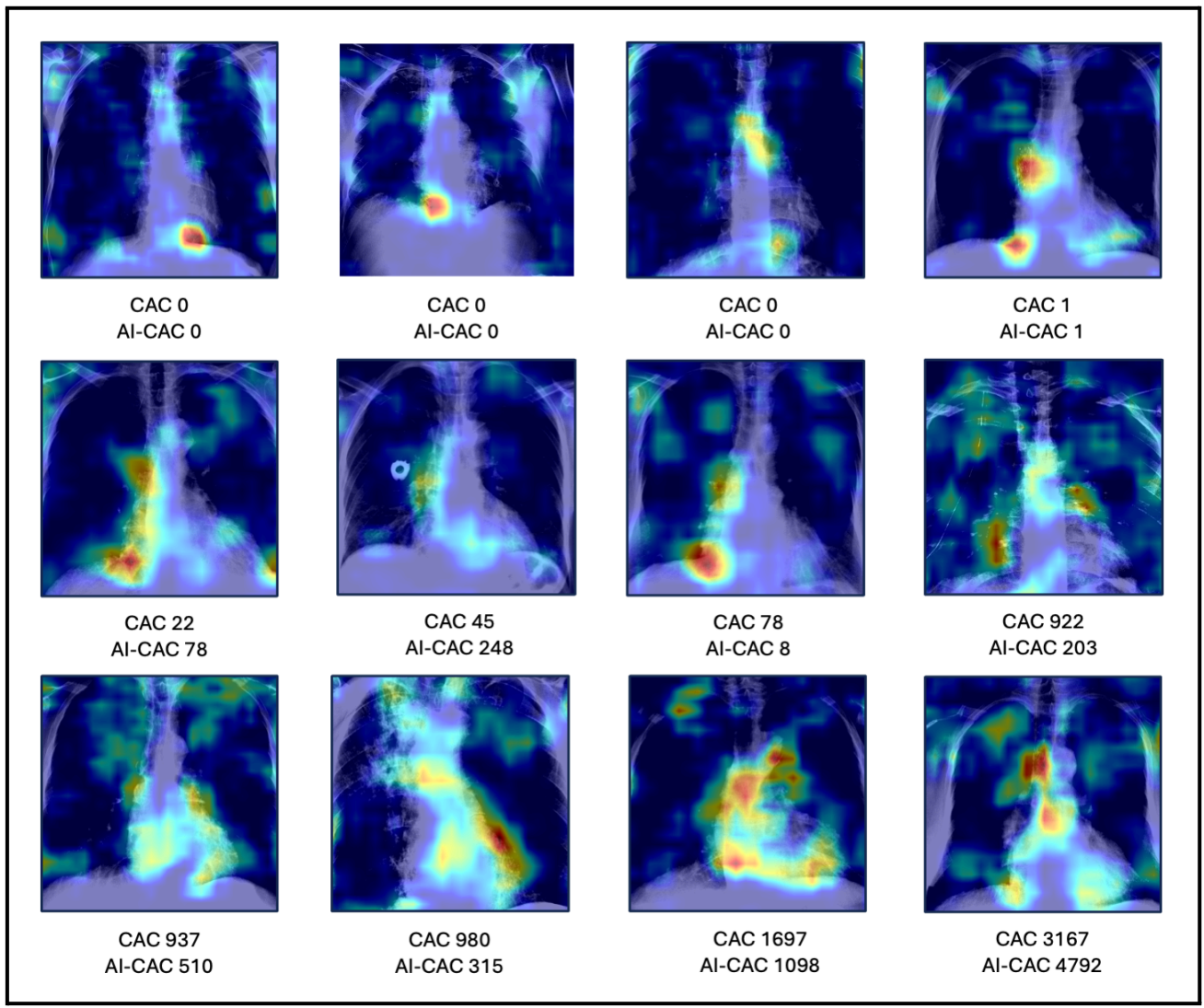

In these importance maps each pixel is represented with a color scale degree reflecting the pixel features importance in the final AI-CAC value output for a given patient. Several representative maps are reported, of patients with different degree of CAC. The most colored pixels in each map represent the anatomical structures with which the model relied on for that specific patient to elaborate the AI-CAC value. Overall, among anatomical structures, the cardiac silhouette, and particularly the coronary course, and the thoracic aorta seem to inform the AI-CAC model.

# SUPPLEMENTARY REFERENCES